\documentclass{article}

\usepackage{microtype}
\usepackage{graphicx}
\usepackage{subcaption}

\usepackage{booktabs} 

\usepackage{epsfig}
\usepackage{amsmath}
\usepackage{amsfonts}
\usepackage{amssymb}
\usepackage{placeins}
\usepackage{url}
\usepackage[normalem]{ulem}

\usepackage{hyperref}

\usepackage[accepted]{icml2021arxiv}

\icmltitlerunning{Synthesizing Irreproducibility}

\usepackage{tikz}
\usetikzlibrary{shapes.geometric}

\usepackage{syn-diagrams}

\begin{document}

\twocolumn[
\icmltitle{Synthesizing Irreproducibility in Deep Networks}




\begin{icmlauthorlist}
\icmlauthor{Robert R. Snapp}{goo}
\icmlauthor{Gil I. Shamir}{goo}
\end{icmlauthorlist}

\icmlaffiliation{goo}{Google, 6425 Penn Ave. Suite 700, Pittsburgh, PA, USA}

\icmlcorrespondingauthor{Robert R. Snapp}{snapp@google.com}
\icmlcorrespondingauthor{Gil I. Shamir}{gshamir@ieee.org}

\icmlkeywords{deep networks, irreproducibility, underspecifiation, nondeterminism}

\vskip 0.3in
]




\printAffiliationsAndNotice 


\newcommand{\be}{\begin{equation}}
\newcommand{\ee}{\end{equation}}

\newcommand{\bea}{\begin{eqnarray}}
\newcommand{\eea}{\end{eqnarray}}

\newcommand{\bi}{\begin{itemize}}
\newcommand{\ei}{\end{itemize}}

\newcommand{\ben}{\begin{enumerate}}
\newcommand{\een}{\end{enumerate}}

\newcommand{\bef}{\begin{figure}[htbp]}
\newcommand{\enf}{\end{figure}}

\newcommand{\bt}{\begin{tabular}{lcllcl}}
\newcommand{\et}{\end{tabular}}

\newcommand{\bd}{\begin{description}}
\newcommand{\ed}{\end{description}}

\newtheorem{theorem}{Theorem}
\newtheorem{lemma}{Lemma}[section]
\newtheorem{corollary}{Corollary}
\newtheorem{proposition}{Proposition}
\newtheorem{condition}{Condition}

\newcounter{example}
\renewcommand{\theexample}{\arabic{example}}
\newenvironment{example}
 {\refstepcounter{example}%
  \vspace{.25cm}%
  \noindent%
  {\bf \boldmath Example \arabic{example}}\\%
  \noindent}
 {\hfill $\Box$ }

\renewcommand{\thecondition}{\Alph{condition}}

\newenvironment{proof}[1]
 {\noindent%
 {\bf \boldmath Proof #1:}}
 {\hfill $\Box$  \\}


\newcommand{\ul}{\underline}
\newcommand{\eref}[1]{(\ref{#1})}       


\newcommand{\dfn}{\stackrel{\triangle}{=}}  
\newcommand{\eqex}{\stackrel{\cdot}{=}}     
\newcommand{\half}{\frac{1}{2}}                 

\newcommand{\comb}[2]{\left ( \begin{array}{c}
 {#1} \\
 {#2} \end{array} \right )}
\newcommand{\comba}[2]{\left (
 \raisebox{-4pt}{$\stackrel{\mbox{\large $#1$}}{#2}$} \right ) }

\newcommand{\avec} {{\mathbf a}}
\newcommand{\evec} {{\mathbf e}}
\newcommand{\dvec} {{\mathbf d}}
\newcommand{\vvec} {{\mathbf v}}
\newcommand{\wvec} {{\mathbf w}}
\newcommand{\xvec} {{\mathbf x}}
\newcommand{\yvec} {{\mathbf y}}
\newcommand{\zvec} {{\mathbf z}}
\newcommand{\uvec} {{\mathbf u}}
\newcommand{\tvec} {{\mathbf t}}
\newcommand{\bvec} {{\mathbf b}}
\newcommand{\kvec} {{\mathbf k}}
\newcommand{\nvec} {{\mathbf n}}
\newcommand{\onevec}{{\mathbf 1}}
\newcommand{\zerovec}{{\mathbf 0}}

\newcommand{\pvec}   {\mbox{\boldmath $\theta$}}
\newcommand{\psivec} {\mbox{\boldmath $\psi$}}
\newcommand{\sigvec} {\mbox{\boldmath $\sigma$}}
\newcommand{\phivec} {\mbox{\boldmath $\phi$}}
\newcommand{\vphivec}{\mbox{\boldmath $\varphi$}}
\newcommand{\tauvec} {\mbox{\boldmath $\tau$}}
\newcommand{\etavec} {\mbox{\boldmath $\eta$}}
\newcommand{\xivec}  {\mbox{\boldmath $\xi$}}
\newcommand{\omegavec}{\mbox{\boldmath $\omega$}}
\newcommand{\kappavec}  {\mbox{\boldmath $\kappa$}}

\newcommand{\Vvec} {{\mathbf V}}
\newcommand{\Wvec} {{\mathbf W}}
\newcommand{\Xvec} {{\mathbf X}}
\newcommand{\Yvec} {{\mathbf Y}}
\newcommand{\Uvec} {{\mathbf U}}
\newcommand{\Tvec} {{\mathbf T}}
\newcommand{\Bvec} {{\mathbf B}}
\newcommand{\Agrid}{\mbox{\boldmath $\Omega$}}

\newcommand{\Pvec} {\mbox{\boldmath $\Theta$}}
\newcommand{\Pvece} {\hat{\Pvec}}
\newcommand{\Psivec} {\mbox{\boldmath $\Psi$}}
\newcommand{\Sigvec} {\mbox{\boldmath $\Sigma$}}
\newcommand{\Phivec} {\mbox{\boldmath $\Phi$}}
\newcommand{\Vphivec}{\mbox{\boldmath $\Varphi$}}

\newcommand{\vect}[1]{\mathbf{#1}}

\newcommand{\pvece}{\hat{\pvec}}

\newcommand{\avecspace}{\Lambda}

\newcommand{\Beta}{{\cal B}}
\newcommand{\Mono}{{\cal M}}

\newcommand{\eventA}{{\cal A}}
\newcommand{\eventB}{{\cal B}}
\newcommand{\eventF}{{\cal F}}
\newcommand{\Sset}{{\cal S}}
\newcommand{\Tset}{{\cal T}}

\newcommand{\Xab}{{\cal X}}

\newcommand{\xit}{x_{i,t}}
\newcommand{\xito}{x_{i,t+1}}
\newcommand{\xitm}{x_{i,t-1}}
\newcommand{\yt}{y_t}
\newcommand{\yto}{y_{t+1}}
\newcommand{\ytm}{y_{t-1}}
\newcommand{\wt}{w_t}
\newcommand{\wtm}{w_{t-1}}
\newcommand{\wto}{w_{t+1}}
\newcommand{\wit}{w_{i,t}}
\newcommand{\wito}{w_{i,t+1}}
\newcommand{\witm}{w_{i,t-1}}
\newcommand{\wi}{w_i}
\newcommand{\D}{\Delta}

\newcommand{\pit}{p_{i,t}}
\newcommand{\pipt}{p_{i,t+} \left (\muito \right)}

\newcommand{\sigmoid}{\mbox{Sigma}}

\newcommand{\robert}[1]{\textcolor{red}{#1}}
\newcommand{\gil}[1]{\textcolor{green}{#1}}

\begin{abstract}
The success and superior performance of deep networks is spreading their popularity and use to an increasing number of applications.  Very recent works, however, demonstrate that modern day deep networks suffer from irreproducibility (also referred to as nondeterminism or underspecification).  Two or more models that are identical in architecture, structure, training hyper-parameters, and parameters, and that are trained on exactly the same training data, yield different predictions on individual previously unseen examples. Thus, a model that performs well on controlled test data, may perform in unexpected ways when deployed in the real world, whose data is expected to be similar to the test data.
We study simple synthetic models and data to understand the origins of these problems.  We show that even with a single nonlinearity and for very simple data and models, irreproducibility occurs.  Our study demonstrates the effects of randomness in initialization, training data shuffling window size, and activation functions on prediction irreproducibility, even under very controlled synthetic data.
While, as one would expect, randomness in initialization and in shuffling the training examples exacerbates the phenomenon, we show that model complexity and the choice of nonlinearity also play significant roles in making deep models irreproducible.
\end{abstract}

\section{Introduction}
\label{sec:introduction}
In recent years, systems in every domain of our lives have been moving away from domain specific mathematical solutions towards using deep networks for prediction, classification, and other problems.  Domains, such as image or video understanding, speech processing and recognition, recommendation systems, and many more, now rely heavily on deep models that are trained on large data-sets of legacy examples.  Due to the substantial improvements driven by deep networks' based techniques, using deep models became much more popular than prior art methods, and in many of these problems domain specific knowledge has also been integrated into the deep models.

Very recently, however, despite the performance accuracy superiority of deep models, results that question other aspects of deep models have started surfacing.  Bias towards the training examples has obviously been one of these aspects.  However, another aspect that has had very limited coverage is that of \emph{irreproducibility} in deep models (also \emph{nondeterminism} or \emph{underspecification} due to over-parameterization).  Normally, deep models have better average prediction accuracy (or objective loss) on validation data. However, their predictions on individual, yet unseen, examples may diverge largely between two separately trained instances of the same model, even if they were defined identically (architecture, parameters, hyper-parameters, optimizers, etc.)\ and were trained on the exact same training data-set. 
Observed \emph{Prediction Differences (PDs)} can become substantial fractions of the actual predictions themselves (see, e.g., \citet{chen20, dusenberry20}).

While focus on irreproducibility has been limited, very recent papers began identifying the problem and its implications.  Recent work \cite{damour20} identified that while various models can appear to perform well on carefully crafted test data, their performance may be unacceptable on data slices of interest when models are deployed to real data.  Because of overparameterization of deep networks, and underspecification of these parameters by the training data set, a trained model converges to a solution it prefers on the training data, which may not be the solution one actually desires for their data slice of interest on the actual real data.  Many applications can tolerate irreproducibilty or deviation between performance on test data and real data.  However, for some applications such model behavior can be detrimental.  For example, two different medical diagnoses for a medical concern may have life-threatening implications.  In reinforcement learning or online Click-Through-Rate (CTR) prediction systems \citep{mcmahan13}, decisions may determine subsequent training examples, and divergence may occur if two models generate different predictions leading to different decisions.

Irreproducibility in deep networks is different from related problems widely studied in the literature.  It is not overfitting.  In the latter, one would observe degradation in performance accuracy due to noise present in the training set mistakenly assumed to be signal. No such degradation is observed with irreproducibility. The relation to prediction uncertainty is more subtle, and harder to distinguish.  Irreproducibility can be considered as a form of uncertainty, but different from the widely studied epistemic uncertainty.  Randomness (or shuffling) of training examples does induce irreproducibility due to two reasons, different trajectories towards the optimum, or a different optimum altogether.  The former is due to epistemic uncertainty, and may be observed as irreproducibility (as we show later), while the latter is caused by multiple identical optima.  Epistemic uncertainty diminishes with more training examples, while irreproducibility due to multiple optima does not.

For convex (linear) models, randomness in training may make optimizers approach the optimum from different directions, thus generating prediction differences due to \emph{early stopping}, where the model prediction is far from the ``true'' optimum as function of the number of training examples seen.  The objective for deep models, on the other hand, will have multiple optima, and many which have roughly equal loss in average over all test examples, but differ on the individual predictions they provide to an example.  For such models, nondeterminism in training may lead optimizers to different optima
\citep{summers21} (see also \citet{nagarajan18}), that depend on the training randomness
\citep{achille17, bengio09}.

Unfortunately, in modern large-scale systems, nondeterminism is unavoidable as highly distributed systems are required to train models with vast amounts of data.  Multiple factors contribute to nondeterminism resulting in model irreproducibility: randomness in model initialization, desirable or undesirable shuffling of training examples, rounding errors in optimizers, the actual training, test, and real data-sets, the model complexity,
the choices of the optimizer, architecture, and hyper-parameters.  While augmentation of data in training and stochastic regularization randomly applied in training \cite{summers21} also influence irreproducibility, we do not consider these here, as they are in a different category of directly and intentionally adding randomness.  Recently, \citet{shamir20s} showed that the choice of activation in a deep network also plays a significant role in exacerbating irreproducibility, where specifically the celebrated \emph{Rectified Linear Unit (ReLU)} \citep{nair10} can be a major contributor.

{\bf Our Contributions:}
We empirically study very simple models on data we synthesize to demonstrate that irreproducibility is indeed a concern even with the smallest simple possible models.  Generating synthetic data and measuring the phenomenon allows us to show that one should expect this problem in deep models due to their non-convex objective surface.  We demonstrate that different initializations of supposedly identical models do lead models to different optima, which elevate prediction differences.  We demonstrate how the effect of randomly shuffling the training examples also increases prediction differences.  We show, however, that even with minimal shuffling, non-linearity exacerbates prediction differences, starting even with non-linearity generated by hidden layers in deep networks with identity activation. Moving, however, towards non-smooth ReLU substantially worsens the effect, where a smooth activation, such as \emph{SmeLU} \citep{shamir20s} or \emph{Swish} \citep{ramachandran17}, exhibits prediction difference larger than identity, but smaller than ReLU.  We show that the PD benefits of a smooth activation are limited locally and diminish with more aggressive shuffling because they arise from fewer local optima.  Too aggressive shuffling diminishes the effects of fewer optima. Interestingly, prediction differences, although negligible, are observed even with convex models with identical initialization and no shuffling.  This is because modern training systems, as TensorFlow, use mini-batches of examples often processed in random order.  Rounding errors in computation accumulate to produce nonzero prediction differences.  These, however, are no longer negligible with deep models with nonlinear activation, even without data shuffling and with identical initialization.  With aggressive shuffling, even convex models can show a high level of irreproducibility stemming from early stopping with a different trajectory to the optimum, as mentioned above.
Finally,
our results show that PDs increase with model complexity, are also affected by the amount of smoothness of a smooth activation, by the choice of optimizer, by choices of other hyper-parameters, and can be mitigated with warm starting models even when aggressive data shuffling is present.

We build the paper starting from the simplest possible case, where our synthetic data is generated from a simple linear model with a few true binary features.  Our baseline trained model is a simple linear model with those binary features.  With aggressive shuffling, we observe elevated epistemic uncertainty (early stopping) driven PDs even with this baseline, but without affecting the prediction loss. 
We first consider a model with the input binary features feeding into a single non-linearity (identity, and then ReLU).  Such a model is misspecified relative to the true data model, and is unable to find the best solution.  However, even with identity activation, it is no longer convex, and the model can find two different solutions due to its overparameterization, in turn increasing PDs.  Accuracy (or prediction loss) is equal for the different models. A ReLU unit can lock the hidden unit to allow only one sign of inputs, again, misspecifying the solution, and requiring more units to explain data with opposite signs. We next move to two units, and demonstrate that while accuracy can improve, PDs actually increase with more degrees of freedom: overparameterization (or underspecification of the model in training).  We continue by looking at deeper models with more hidden units, showing that PDs increase.  We then demonstrate similar behavior for models in which \emph{embeddings} replace the binary inputs.  We conclude by showing similar behavior for deep models where the data is generated by a non-linear (quadratic) model.  Here, the deep models substantially outperform a linear baseline in terms of prediction loss, but still exhibit irreproducibility as seen for the other models.

{\bf Related Work:}
While there is ample literature about uncertainty in deep models, as mentioned, irreproducibility has been given very little attention.  Ensembles \citep{dietterich00} that reduce uncertainty \citep{lakshminarayanan17} can also reduce PD.  They do, however, impose more system complexity and technical debt, and they can trade model accuracy for better reproducibility, especially in large scale systems which operate in the underfitting regime.  This is because if one is constrained by computation (number of flops per single example), in order to keep the constraint, the components of the ensemble must consist of narrower layers than those of the single network which is compared to.  In the underfitting regime, where adding units still improves accuracy, this degrades model accuracy.  Ensembling the components improves it by less than co-training all parameters in a single network.

\emph{Distillation} \citep{hinton15} is becoming a popular method to transfer information from a high complexity expensive teacher model to a simpler student model, which is trained to learn from the teacher.  The simple model can be deployed, saving on deployment resources.
\emph{Co-distillation}, proposed by \citet{anil18} (see also \cite{zhang18}),
addresses irreproducibility by symmetrically distilling among training models in a set, pushing them to agree on a solution.  Only a single model from the set, which is more reproducible, must be deployed. The additional constraints, however, can impair the accuracy of the deployed model.  \emph{Anti-Distillation} \cite{shamir20a} embraces ensembles, and adds a loss that forces components to diverge from one another, together capturing larger diversity of the solution space, giving more reproducible ensembles.  Additional approaches to address irreproducibility anchor the trained model to some constraints forcing it to prefer solutions that satisfy the constraints over others \cite{bho21, shamir18}.  \citet{damour20, summers21} recently studied the irreproducibitily problem on benchmark data-sets.

\section{Set Up and Evaluation Metrics}
\label{sec:setup}
To train and evaluate models on synthetic data we built a simulation framework that consists of data generation, a model training pipeline and an evaluation component.

{\bf Data Generation:} Experiments described are on data that was generated by a true linear model (Sections~\ref{sec:single}-\ref{sec:linear}) and a true quadratic model (Section~\ref{sec:quad}).  The linear model is described here, and the quadratic in Section~\ref{sec:quad}.  The linear model is a sparse model with $d=32$ binary features.  Training and test example $\xvec_t \in \{0, 1\}^d$ is a sparse vector whose components take value $1$ for features present in the example, and $0$ for features that are not present.  We usually partition the $32$ components into two sets of $16$, in each set, dimension $j = 1,2,\ldots,16;$ takes value $1$ with i.i.d.\ probability $6/(j\pi)^2$.  This guarantees nonuniform probability for different features, with some ``long tail'' of the higher indexed features, where low-indexed features appear more frequently (e.g., for $j=1$, with probability $0.61$).

True log-odds weights $\theta \in \mathbb{R}^d$ for each of the $32$ features are drawn randomly once.  To diversify values of $\theta$, for each value, one of $3$ normal distributions is picked with uniform probability, and then $\theta_j,~j=1,2,\ldots,32;$ is drawn from that distribution.  The distributions are $\mathcal{N}(\mu, \sigma^2 = 1)$, with $\mu \in \{-2, 0, 2\}$.  The label $y_t \in \{-1,1\}$ for training example $t$ is drawn with probability computed by the Sigmoid of the cumulative log-odds, given by
\be
 \label{eq:true_example_prob}
 p_{\theta} (y_t|\xvec) = \sigma \left (y_t \xvec^\mathcal{T} \theta \right ) \dfn
 \frac{1}{1+\exp(-y_t \xvec^\mathcal{T} \theta ) }
\ee
where $\mathcal{T}$ is the transpose operator.  The label $y_t$ is drawn from a Bernoulli distribution with probabilities in \eref{eq:true_example_prob}.

{\bf Training:} Each model trains with a single pass over $T=2^{26} \approx 65M$ examples generated as described, attempting to learn corresponding parameters to explain the observed labels.  For deep models, the model inputs are the vectors $\xvec_t,~t=1,2,\ldots,T$.  For the baseline linear model, the model learns weights $\wvec \in \mathbb{R}^d$ corresponding to the dimensions of $\theta$.  Each experimental setting is trained and evaluated $32$ times, consisting of $M=16$ pairs of models. A pair of models trains on the same sequence of example/label pairs $\{\xvec_t, y_t \}_1^T$.  If an experiment uses an identical initialization, the parameters of the models in the pair are initialized identically, as well.  To remove dependence on a specific sequence $\{\xvec_t, y_t \}_1^T$ (and specific initialization values in experiments with identical initialization), each pair of models has its own example/label pair sequence (and its own initialization values), which are equal for the two models in the pair, but not to those of other pairs.

Using the TensorFlow Keras framework, training is done in mini-batches of $s=32$ examples.  Randomness in updates can occur within a mini-batch as in realistic systems, where one has no control of order of operations inside a mini-batch.  To test effects of shuffling, we define a \emph{window size} $z$ of the number of mini-batches over which shuffling can occur.  For $z=1$ (or $\log z = 0$), shuffling of training data occurs only inside the mini-batch.  With $z > 1$, the training data of a pair of models is partitioned into windows of size $zs$, and training examples are randomly shuffled inside each window differently for each of the two models in the pair.

To illustrate effects of the choice of optimizer and hyper-parameters,
we considered two different optimizers, AdaGrad \cite{duchi11} (per-coordinate learning schedule) with different learning rates (1.0, 0.1, with accumulator initialization of 0.1), and SGD with learning rates 0.1 and 0.01, decay rate 0.001, and momentum 0.9.  Optimizers minimize binary logistic (cross-entropy) loss on the training data.

{\bf Evaluation and Metrics:}
For each pair of models, an identical evaluation set of $N=2^{16}=16K$ examples was drawn from the data model defined above.  An accuracy metric and a reproducibility metric were generated and averaged over the $16$ pairs.  The accuracy of a single model on the evaluation set was measured in terms of \emph{average excess label loss} given for model $m=1,2,\ldots, M$ by
\be
\label{eq:excess_loss}
L(m) \dfn \frac{1}{N} \sum_{n=1}^N \sum_{y \in \{-1,1\}} 
p_{\theta}(y|\xvec'_n) \log \frac{p_{\theta}(y|\xvec'_n)}{\hat{p}_m (y | \xvec'_n )}
\ee
where $\xvec'_n$ denotes the $n$th evaluation example for the model, $p_{\theta}(y|\xvec'_n)$ is the true probability of $y$ for this example \eref{eq:true_example_prob},
and $\hat{p}_m (y | \xvec'_n )$ is predicted for $y$ by model $m$.  The quantity in \eref{eq:excess_loss} has a notion of expected \emph{regret} on the evaluation set. It measures the extra average loss a model incurs over an empirical entropy rate of the evaluation set.  (Clearly, other forms of metrics could be used here, but they are expected to demonstrate similar qualitative behavior.)

Various metrics can be used to measure prediction difference (PD) \cite{chen20, shamir20a, shamir20s}. They all, however, demonstrate similar qualitative behaviors.  Here, we choose a related, but slightly different metric of relative PD on the positive label
\be
\label{eq:pd}
\D_r = \frac{1}{N}\sum_{n=1}^N \frac{2}{M} \sum_{k=1}^{M/2}
\frac{\left | \hat{p}_{2k}(y=1|\xvec'_n) - \hat{p}_{2k-1}(y=1|\xvec'_n) \right |}
{p_{\theta}(y=1|\xvec'_n)}
\ee
where in this definition we assumed that for any $k$, the models $m \in \{2k-1, 2k\}$ are a pair of models.  This metric deviates from the definition of PD in \cite{shamir20s} by averaging over the pair PDs instead of over PDs relative to the expectation on the $M$ models.  This is because in our experiments we have cases in which different pairs have different conditions, averaging out the effects of, for example, a specific initialization vector.  Relative PD is measured as fraction of the true probability of the positive label. Relative PD, on one hand, gives a metric which reduces dependence on the value of this probability, but on the other, does exacerbate the effect of small predictions.  However, with our data model set up the latter does not appear significant.  Since we know the true label probability, our ratio is normalized by its value, unlike other works, that had no knowledge of this value and used expectation of predictions instead.  This reduces variances
of our measurements.

\section{Linear Data Model with Single and Double Hidden Units}
\label{sec:single}
\begin{figure}[h]
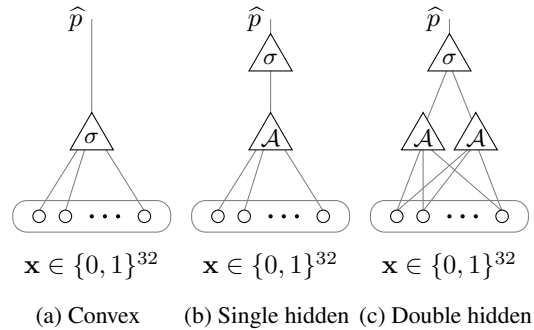

    \centering
    \subcaptionbox{Convex\label{convex}}{\flatLinearModel}
    \subcaptionbox{Single hidden\label{single}}{\singleHiddenUnit}
    \subcaptionbox{Double hidden\label{double}}{\doubleHiddenUnit}
    \caption{Simple models.  Left: baseline convex linear model. Middle: Single activation network. Right: Double activation network.}
    \label{fig:elementary_models}
\end{figure}

We start with the baseline model, and with simple models with a single hidden layer, with one or two units.  Fig.~\ref{fig:elementary_models} shows generic graphs of the three simplest models. Binary inputs enter at the bottom.  Each link on the graphs is associated with a learned weight $w^\ell_{i,j}$, connecting at layer $\ell$ between the $j$th element in the layer below the link (input) and the $i$th element in the layer above the link (output).  Triangles with $\sigma$ designate the Sigmoid function as in \eref{eq:true_example_prob}, and triangles with $\mathcal{A}$ designate a generic activation that could be identity, ReLU, SmeLU, or any other non-linearity.  Before applying the non-linearity, in both cases, a learned bias $b^\ell$ is added to the value forward propagating from the layer below.  Thus the predictions for label $y \in \{-1,1\}$ of the three models in Fig.~\ref{fig:elementary_models} are defined as
\bea
 \label{eq:convex_linear}
 \hat{p}_{\text{linear}} &=& \sigma \left [ y \cdot (b^0 + \xvec^{\mathcal{T}}\wvec^0) 
 \right ]\\
 \label{eq:single}
 \hat{p}_{\text{single}} &=& \sigma \left \{ y \cdot \left [b^1 + w^1_1 \cdot \mathcal{A} (b^0 + \xvec^{\mathcal{T}}\wvec^0) \right ] \right \}\\
 \label{eq:double}
 \hat{p}_{\text{double}} &=& \sigma \left \{ y \cdot \left [b^1 + \wvec^{1\mathcal{T}} \cdot \mathcal{A} (\bvec^0 + \Wvec^0 \xvec) \right ] \right \}
\eea
where for the double hidden units network, $\Wvec^1$ is a matrix of dimensions $2 \times 32$.

\begin{figure}
    \centering
    {\includegraphics[width=8cm]{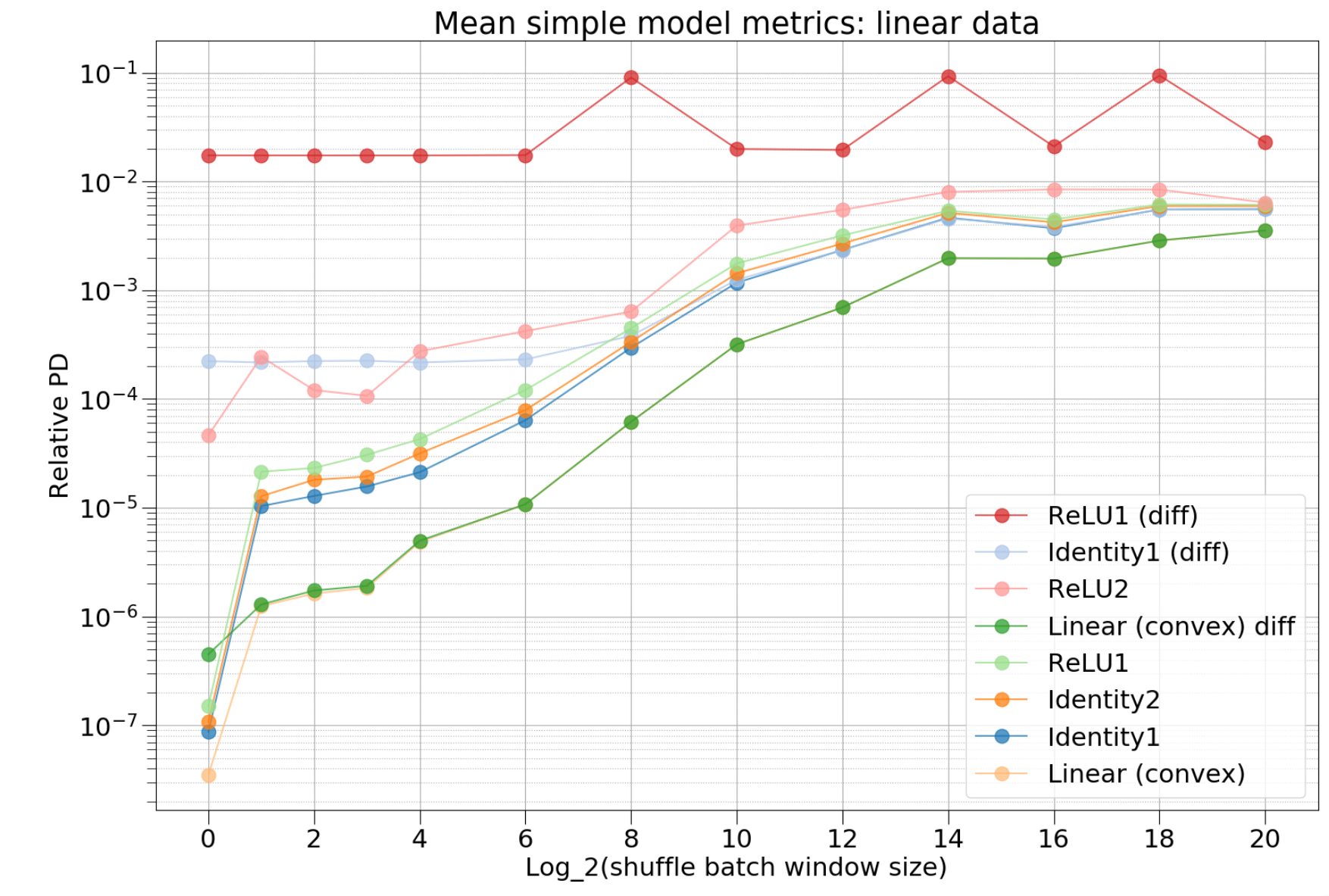}}
    {\includegraphics[width=8cm]{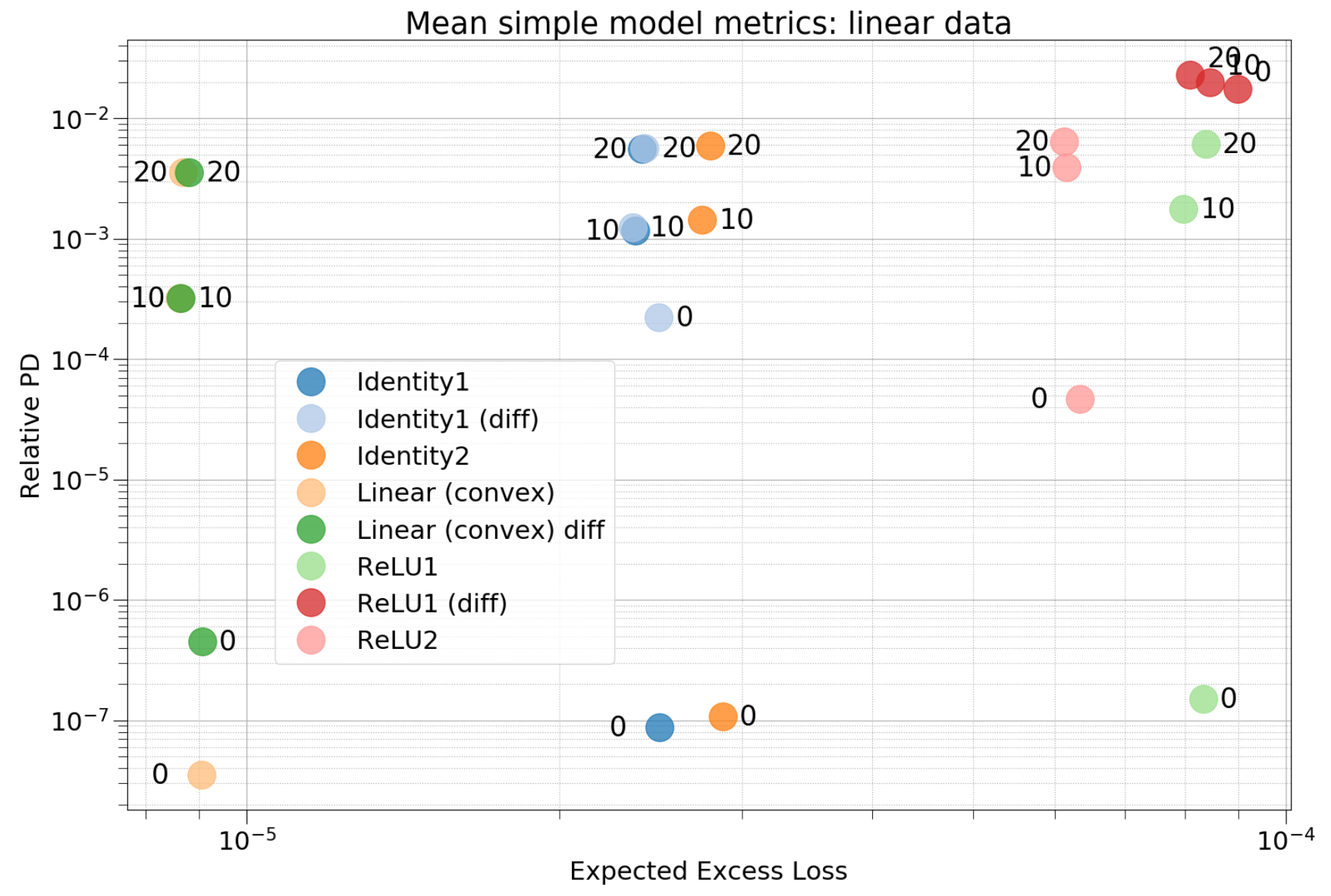}}
    \caption{Mean metrics for various simple models, all trained with AdaGrad with learning rate = 0.1. 
    Top: Relative PD $\D_r$ as function of logarithm of shuffling window size $\log_2 z$.
    Bottom: $\D_r$ as function of expected excess loss $L$ for various shuffling sizes (with $\log_2 z$ labeled for each point).
    Activation $\mathcal{A}$ is labeled in the legend for each curve, followed by the number of hidden units. Models labeled with ``diff'' were initialized with distinct initial conditions.}
    \label{fig:simple_model_metrics}
\end{figure}

Fig.~\ref{fig:simple_model_metrics} shows expected relative PD over the $16$ model pairs as function of the shuffling window size (top) and as function of the excess loss (bottom) for $3$ values of shuffling window size $z \in 2^{\{0, 10, 20\}}$.  Curves are shown for identical initializations of the parameters in a pair of models (link weights and biases), and for distinct initialization of each model in the pair (marked as ``diff'' in the legend). Curves are shown for the baseline convex linear model with prediction in \eref{eq:convex_linear}, models with a single hidden unit  \eref{eq:single}, with either $\mathcal{A}$ as identity, or as ReLU, and for models with two hidden units \eref{eq:double}, with both activations.

As we move left on the bottom graph, accuracy improves (loss decreases).  As we move down, PD improves.  Consistent with results on real and benchmark data-sets, e.g., \cite{shamir20s}, while PD changes with identical or distinct initialization and with the shuffling window, loss on evaluation data appears to be equal for a given model regardless of these factors.  The baseline convex linear model, expectedly, exhibits the best PD curve.  Because it is properly specified to the data generation model, where the other models are misspecified, the linear baseline also exhibits the best loss (as shown in Section~\ref{sec:quad}, this is only an artifact of the data generation).  With identical initialization and no shuffling, while the linear baseline has the best PD, other models are not far behind.  As shuffling increases, there is a sustained gap between the linear baseline and the deep models.  With aggressive shuffling, PD of the linear model is also high. Time series evaluations, though, show that, while for the other models there are limited improvements with more training examples, for the linear model, PD improves with more training examples.  This suggests that PD for the linear model is dominated by early stopping, whereas with the deep models there are more degrees of freedom that yield prediction differences.
\begin{figure}
 \centering
 {\includegraphics[width=7.5cm]{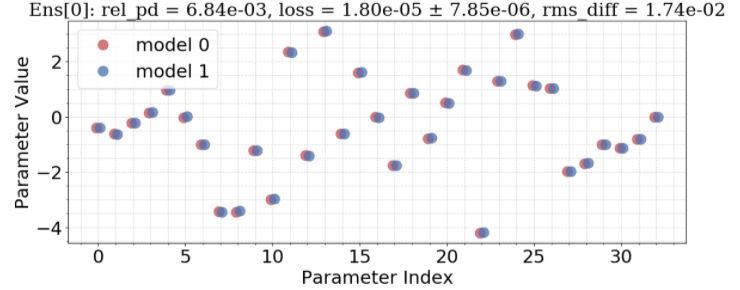}}
 {\includegraphics[width=7.0cm]{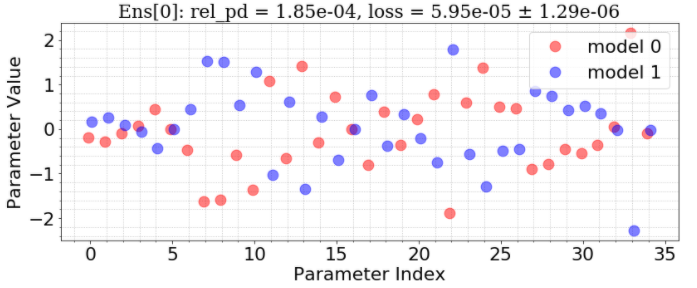}}
 {\includegraphics[width=7.0cm]{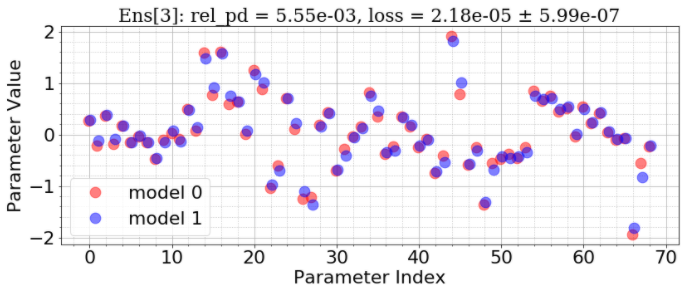}}
 {\includegraphics[width=7.5cm]{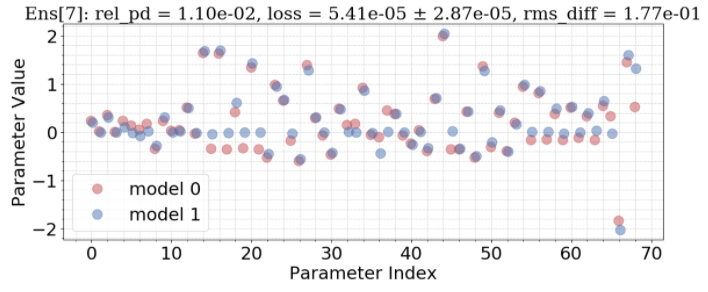}}
 \caption{Trained weights for pairs of models as function of weight index.  From top to bottom: 1. Linear model with distinct (nonidentical) initial conditions and $\log_2 z = 20$;
 2. Identity activation single unit, distinct initial conditions, $\log_2 z = 0$;
 3. Identity activation, two units, same initial conditions, $\log_2 z = 20$;
 4. ReLU, 2 units, same initial conditions, $\log_2 z = 20$.}
\label{fig:simple_model_weights}
\end{figure}
\begin{figure}
\centering
 {\includegraphics[width=7.0cm]{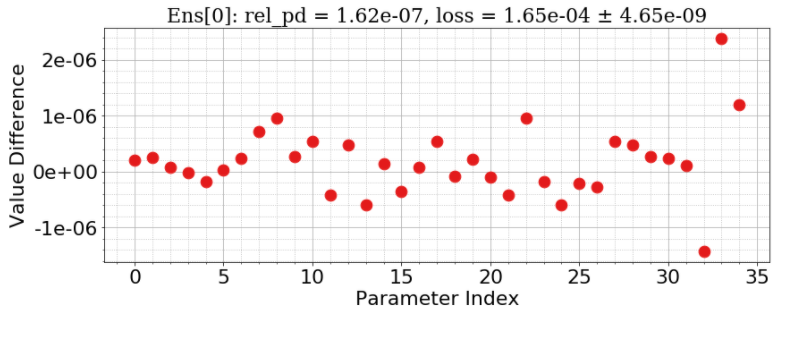}}
 {\includegraphics[width=7.0cm]{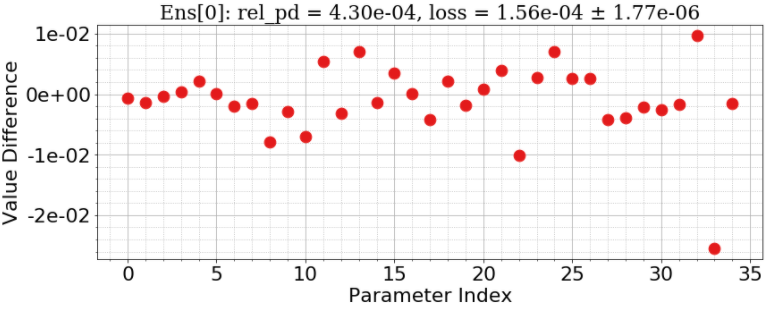}}
 \caption{Differences between corresponding weights for a ReLU1 model pair for
 $\log_2 z = 0$ (top) and $\log_2 z = 20$ (bottom).}
 \label{fig:relu1_deltas}
\end{figure}

With different initializations of models in each pair, we observe the effect of the activation.  While the PD of the convex baseline somewhat degrades with little shuffling (but not with more shuffling), PDs of the deep models degrade much more with little shuffling.  PD for the smooth identity activation with a single mini-batch degrades by orders of magnitude relative to the same activation with identical initialization (from $O(10^{-7})$ to $O(10^{-4})$).  ReLU degrades substantially to $\D_r = O(10^{-2})$, which is retained with more shuffling.  ReLU with two units, even with identical initialization already clearly demonstrates orders of magnitude higher PD than the baseline.  Such degradation is not present with the smooth identity activation with 2 units. 
Enough shuffling with equal initialization leads to PD levels of equal or higher orders to those with different initializations.  The benefits of the smooth identity diminish (and eventually disappear) with more shuffling.

To understand the effects of the different factors, we studied the weights that pairs of trained models converged to.  Fig.~\ref{fig:simple_model_weights} shows examples of weights learned by pairs of models as function of the parameter index.  For each model a different color is used. With the baseline linear model, even with aggressive shuffling and different initialization, the points almost match and seem to deviate very little for all pairs.  Note, however, that if we measure the cosine between the difference vectors $\wvec_{2k} - \theta$, and $\wvec_{2k-1} - \theta$, where the trained vectors $\wvec_{j}$ are those of the $k$th pair (both trained with different initial weights), we observe a cosine of almost $1$ with no shuffling, but one of $0.191$ with maximum shuffling, suggesting that the shuffling results in the model approaching the optimum from a different direction.

For the single activation models, with different initialization, we observe two different phases.  One is similar to the linear model, but the other has weights that converge to opposite signs (with little noise) as shown in the second image in Fig.~\ref{fig:simple_model_weights}.  We were unable to observe the latter with identical initialization pairs, even with the most aggressive shuffling. Single ReLU models appeared slightly more noisy than those with identity activation.  The bottom images in Fig.~\ref{fig:simple_model_weights} for two ReLU units show clearer deviations between the models with minimum shuffling, and stronger deviations with maximum shuffling. These conclusions are even more obvious by observing Fig.~\ref{fig:relu1_deltas}, that shows only the weight differences between models in a pair for a single ReLU model.  The scale of differences with minimum shuffling is $O(10^{-6})$ and increases to $O(10^{-2})$ with maximum shuffling.

\section{Multi Layers}
\label{sec:multi}
\begin{figure}
    \centering
    {\includegraphics[width=8cm]{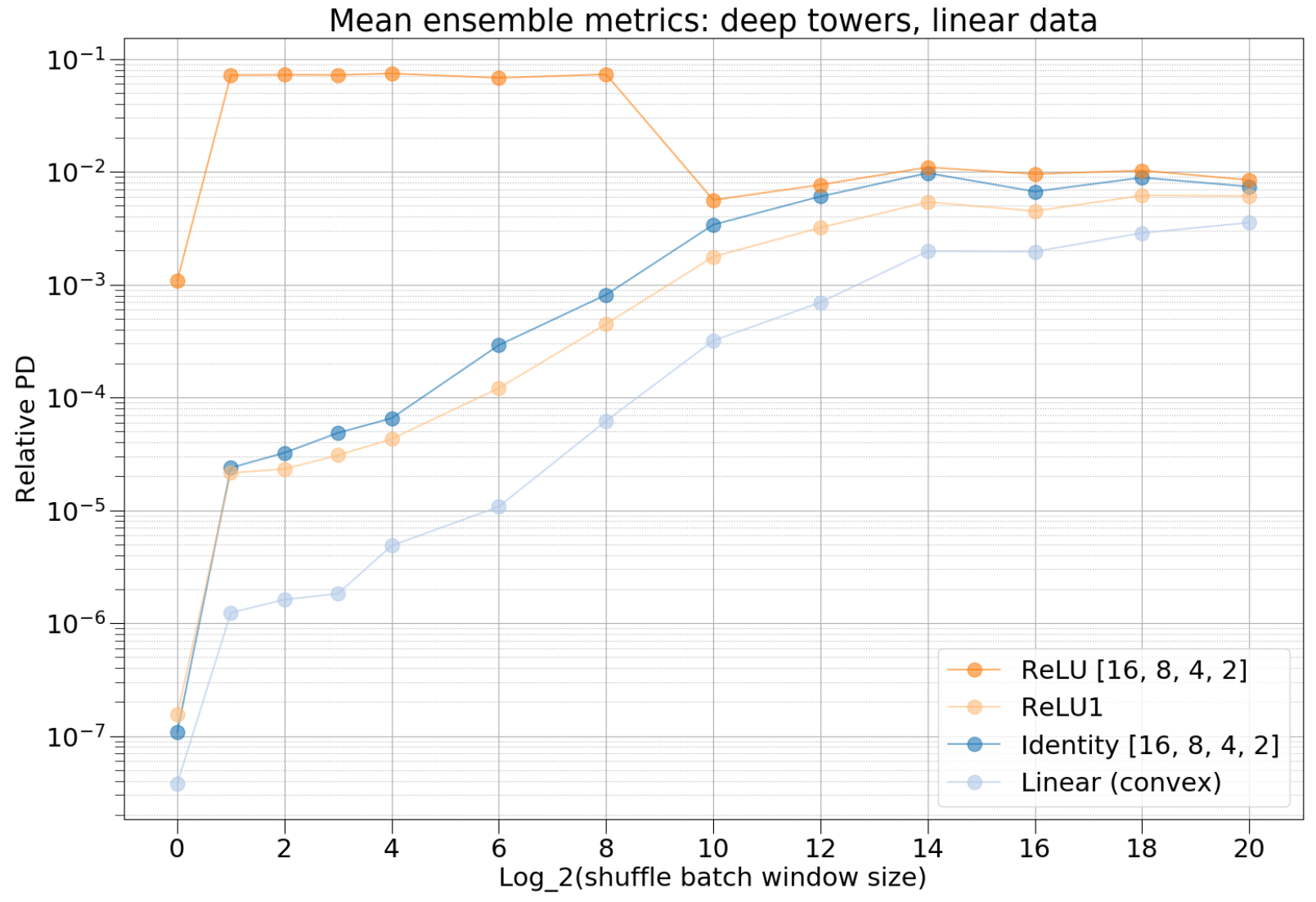}}
    {\includegraphics[width=8cm]{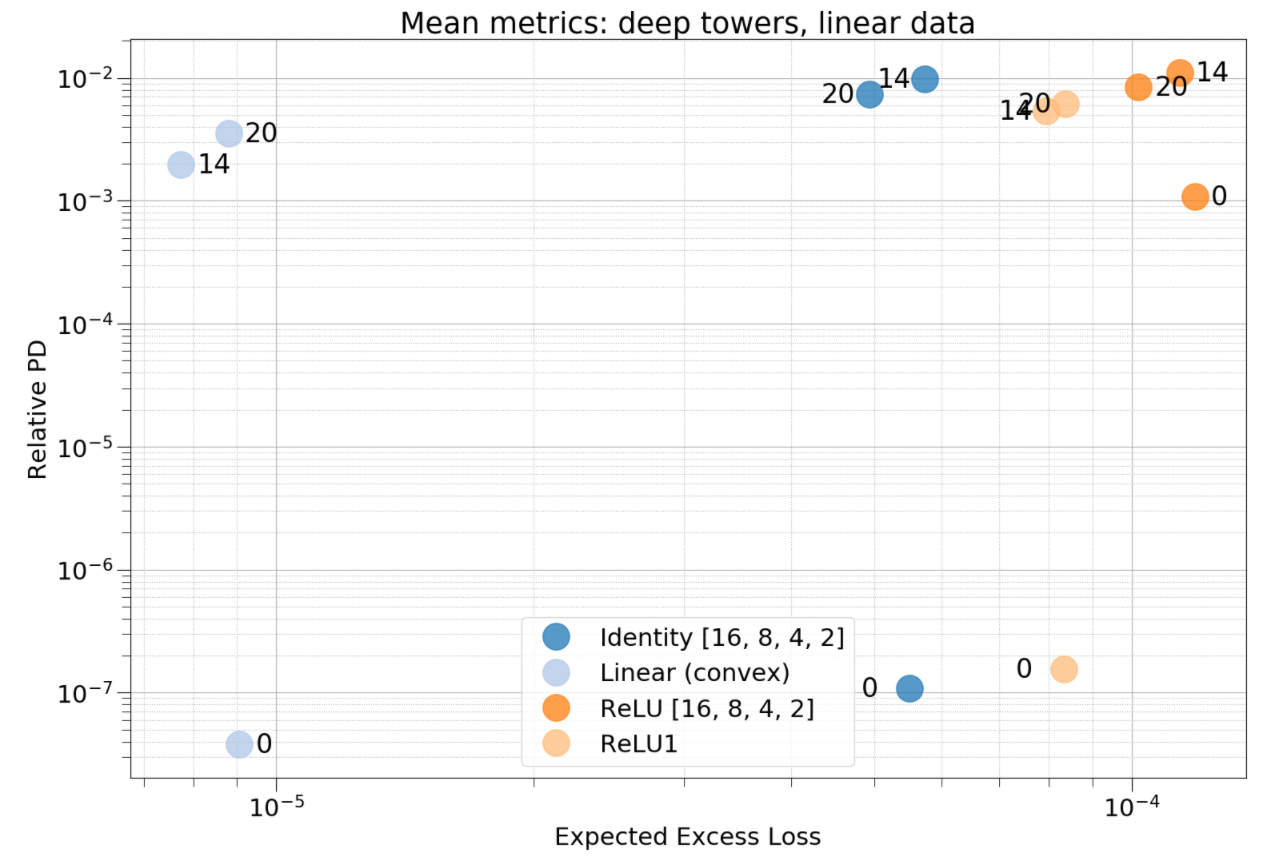}}
    \caption{Ensemble mean metrics for multi hidden units narrow tower models, with hidden layer widths $[16, 8, 4, 2]$, identity and ReLU activation.  Top: $\D_r$ Vs. $\log_2 z$. Bottom: $\D_r$ Vs. $L$, different $\log_2 z$ values.}
    \label{fig:narrow_model_metrics}
\end{figure}

\begin{figure}
\includegraphics[width=7.5cm]{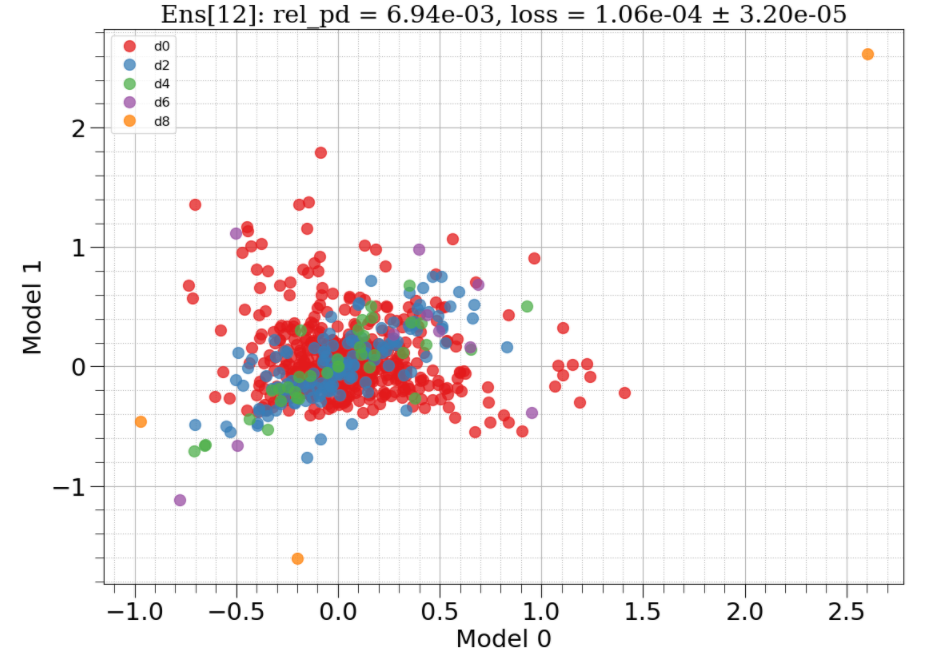}
\caption{Correlation plot of the parameters of a single pair of identically initialized ReLU models with hidden layers of widths $[16, 8, 4, 2]$ comparing values of learned parameters of one model to the other, where each layer is shown with a different color.}
\label{fig:cor_narrow}
\end{figure}

We next move to more complex models with the same input but deeper with more units as shown in Fig.~\ref{fig:narrow_tower_model}.  With the linear data model, these are still misspecified.
As shown in Fig.~\ref{fig:narrow_model_metrics}, this yields worse PD for the ReLU model even with minimal shuffling and with identical initialization of the model pair.  Some model instantiations (even with little shuffling) are stuck, unable to learn, degrading expected loss and PD substantially.  Fig.~\ref{fig:cor_narrow} demonstrates the ``wide correlation cloud'' between the learned weights of two models in an identically initialized pair of ReLU activated models. This cloud suggests that the models converge to very different optima (although, despite that, when they manage to learn, excess loss on evaluation data is similar). PDs of both the ReLU and the identity activation are high and worse than a single ReLU also with more shuffling.  With minimal shuffling, identity still retains low PDs of $O(10^{-6})$.
\begin{figure}
    \centering
    \subcaptionbox{Deep tower\label{fig:narrow_tower_model}}{\multiLayerTower}
    \subcaptionbox{Wide embedding model\label{fig:wide_model}}{\embeddingWide}
    \caption{Deep tower and wide embedding models}
\end{figure}

\section{Wider Layers with Trained Embeddings}
\label{sec:linear}
In this section, we study wider models where inputs are mapped to an embedding space, as shown in Fig.~\ref{fig:wide_model}.  Each feature in each of the two sets of $16$ features is mapped to a two dimensional vector $\evec \in \mathbb{R}^2$.  For each example, the vectors representing all active features in a set of $16$ (nonzero components of $\xvec$) are summed into an embedding input, which is fed into $1000$ hidden units, activated by $\mathcal{A}$.
\begin{figure}
    \centering
{\includegraphics[width=8cm]{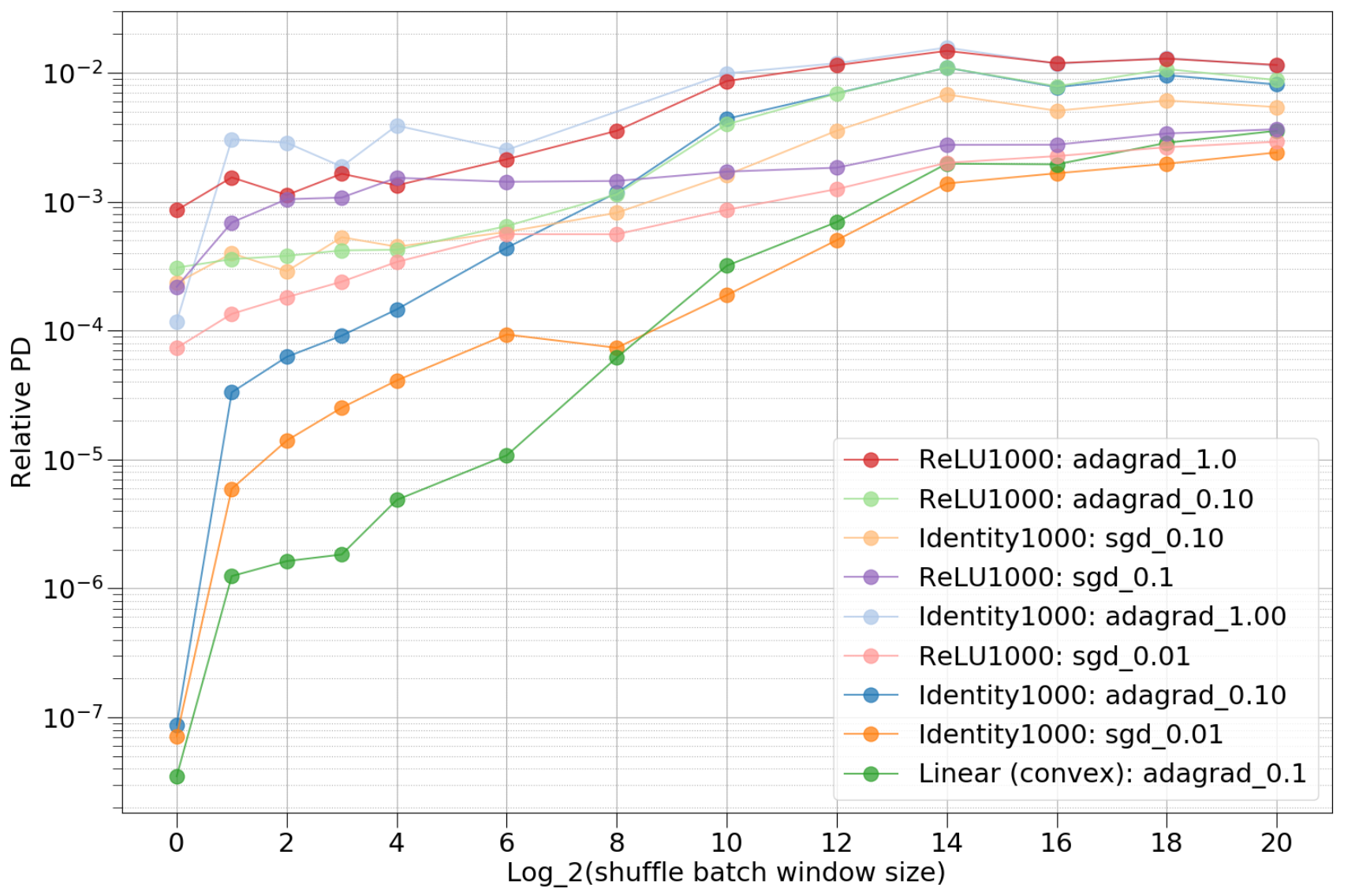}}
{\includegraphics[width=8cm]{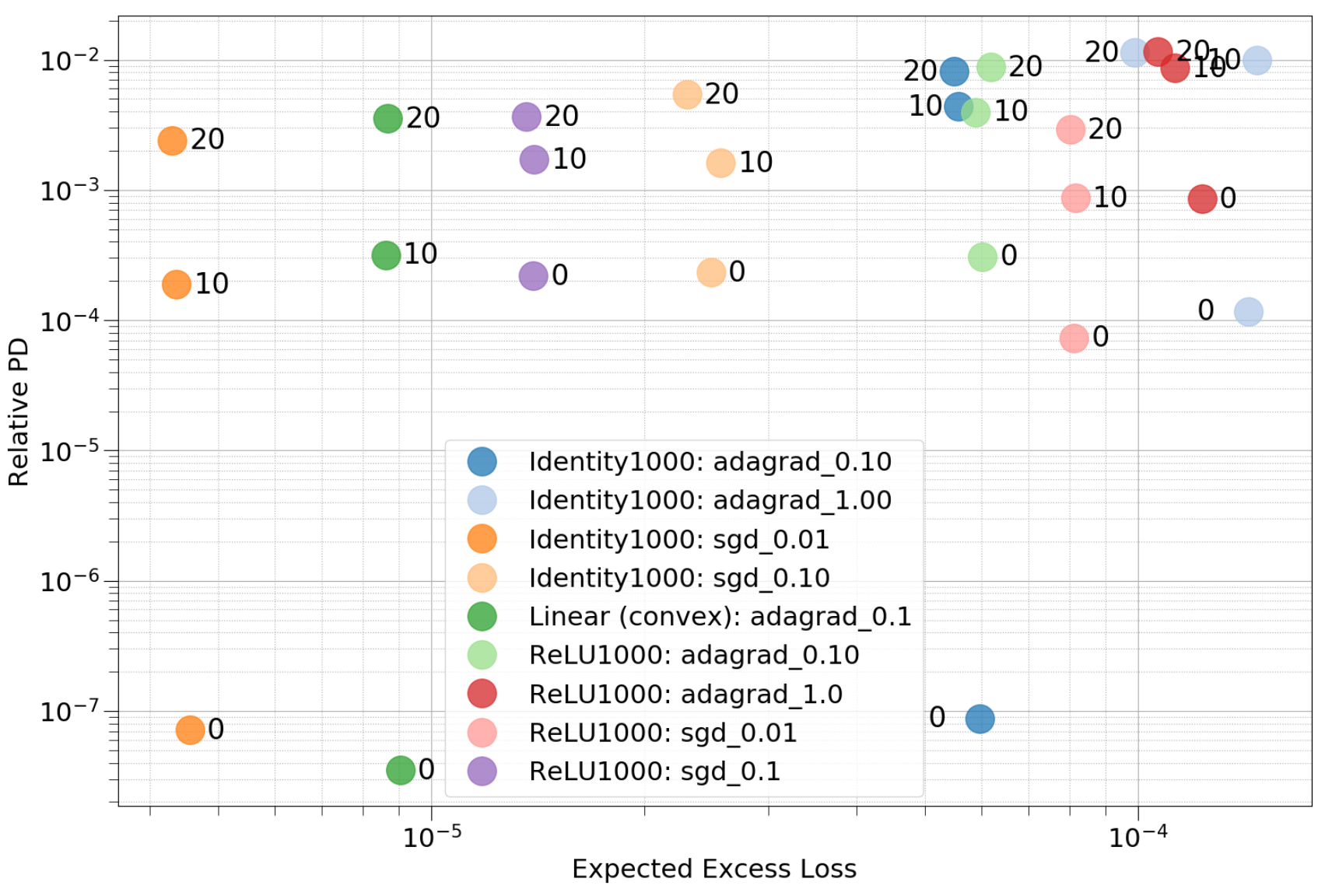}}
    \caption{Ensemble mean metrics for wide models with two two-dimensional embedding inputs, a single $1000$ units hidden layer, for different activations, optimizers and learning-rates.  Top: $\D_r$ Vs. $\log_2 z$. Bottom: $\D_r$ Vs. $L$ for different values of $\log_2 z$.}
    \label{fig:wide_model_metrics}
\end{figure}
\begin{figure}
    \centering
    \includegraphics[width=8cm]{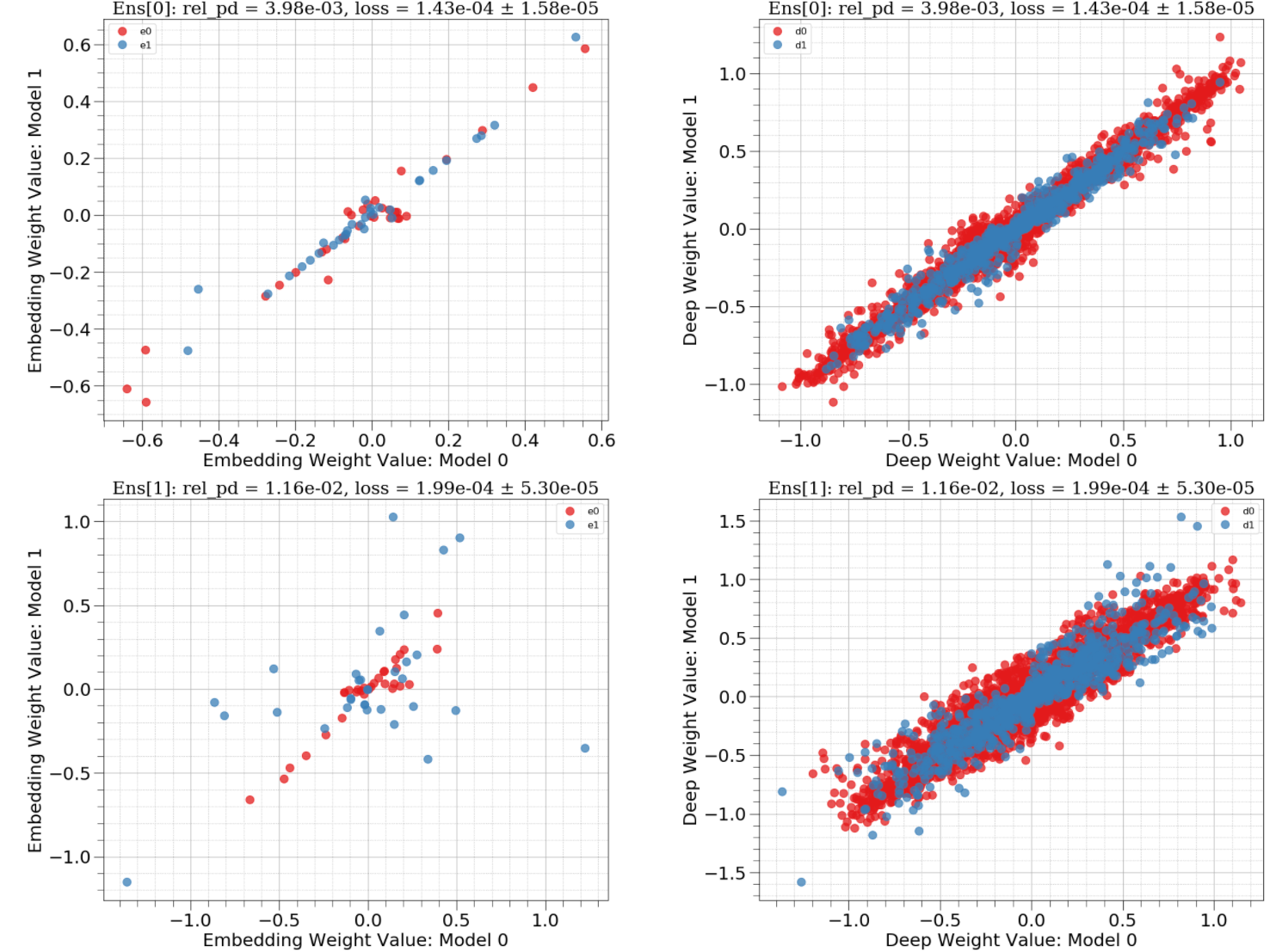}
    \caption{Learned model parameters' correlation graphs of two different pairs of models initialized with common initial conditions, trained with Adagrad with learning rate 0.1 and $\log_2 z = 10$. A row represents a model pair.  Left graphs show correlations of embedding units, and right of layer weights. Red is used for input edges to a unit, and blue for output edges from the unit.}
    \label{fig:wide_model_wc}
\end{figure}

Fig.~\ref{fig:wide_model_metrics} shows PDs as function of the shuffling window and the excess loss for different activation functions, optimizers, and learning rates for model pairs with identical initialization.  Again, we observe mostly consistent excess loss for the same configurations while PDs change with shuffling.  Similar behavior to previous models is observed as function of the activation, with the baseline linear having the best PD curve.  However, loss and PD are also affected by the choice of optimizer and its hyper-parameters, where too large learning rates can degrade loss and PD.  Interestingly, well tuned learning rates even lead to improved loss over the baseline linear with the identity activation.  Again, ReLU exhibits worse PDs even with minimal shuffling.  Fig.~\ref{fig:wide_model_wc} shows correlation curves for two pairs of ReLU models for both embeddings and hidden weights.  If embeddings are aligned, a narrow cloud is observed for the hidden weights.  However, in pairs in which embedding weights are less correlated, hidden weights also deviate more substantially.

\section{A Quadratic Data Model}
\label{sec:quad}
A linear data model may be too simplified for many real cases.  To demonstrate irreproducibility in other data models, we also studied a synthetic quadratic model with $d=32$ inputs dimensions.
Here, the 32 inputs were partitioned into 8 blocks of 4 units, $\vect{x}^T = [\vect{x}_1^T, \ldots, \vect{x}_8^T]$, and the log-odds ratio for the label of each pattern is given by $\vect{x}^T \Theta \vect{x}$, where $\Theta = \mathrm{diag}[L_1, \ldots, L_8]$.
Each $L_i$ is a lower-triangular $4\times 4$ matrix, whose 10 components are generated independently from a mixture of three normal distributions, as before. In each block of $\vect{x}$, the $j$-th feature is selected with a prior probability $90/(j\pi)^4$, for $j = 1, 2, \ldots, 4$, to obtain a comparable tail distribution. Models, containing two hidden layers, with sizes $[1024, 512]$, were trained as before, with a variety of activation functions, including Swish ($y = x \sigma (\beta x)$), and SmeLU ($y = (x+\beta)^2/(4\beta)$ for $|x| < \beta$, $0$ for $x \leq -\beta$, and $x$ for $x \geq \beta$), with different $\beta$ parameters.

\begin{figure}
    \centering
{\includegraphics[width=8cm]{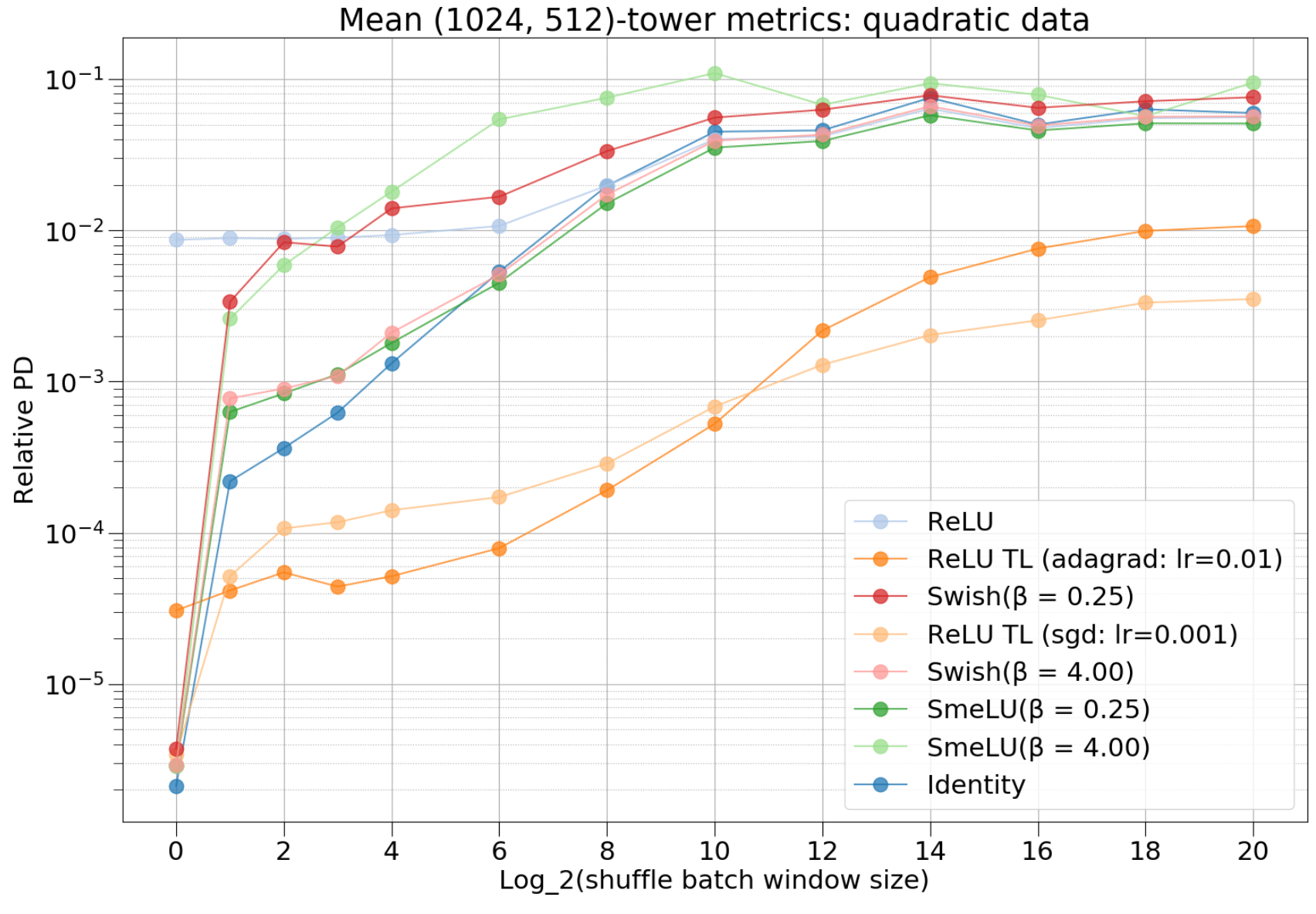}}
{\includegraphics[width=8cm]{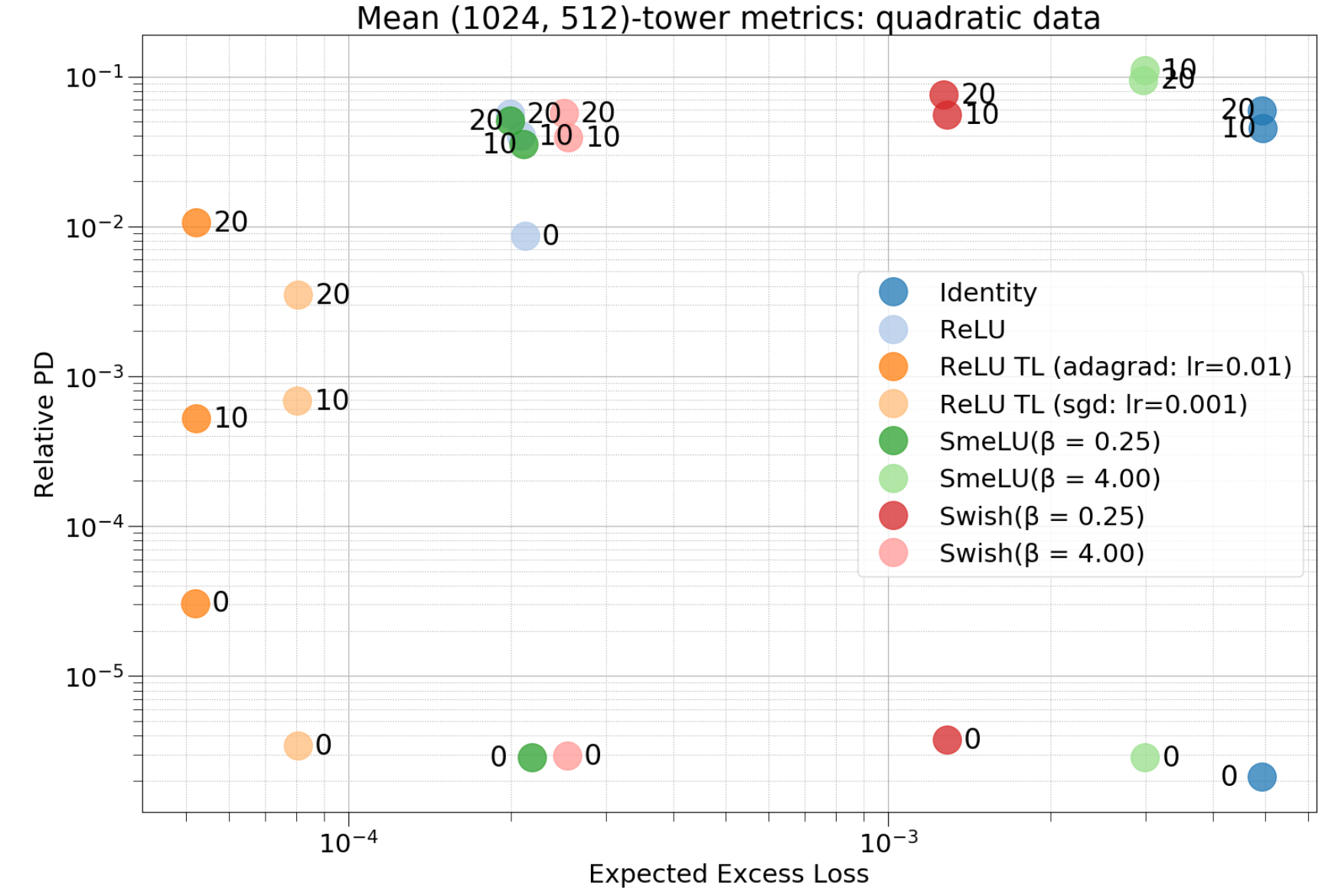}}
    \caption{Ensemble mean performance metrics for two hidden layers $[1024, 512)]$ models with different activations and a quadratic data generation model.  Models labeled \emph{TL} are initialized using warm start transfer learning. Top: $\D_r$ Vs. $\log_2 z$.
    Bottom: $\D_r$ Vs. excess loss $L$ for different values of $\log_2 z$.}
    \label{fig:quadratic_towers}
\end{figure}

Fig.~\ref{fig:quadratic_towers} shows PD and excess loss as function of shuffling for identically initialized model pairs.
The graphs are very similar to the linear model, with the exception of superior loss with non-linear activations, with very poor performance for the baseline linear model.
Again, excess loss seems unaffected by shuffling, whereas PD in
Fig.~\ref{fig:quadratic_towers} also shows how properly tuned SmeLU and Swish can improve PD (with comparable loss) over ReLU as long as shuffling is limited. As shuffling becomes more aggressive, PD gains diminish and eventually disappear because aggressive shuffling can still find different optima even when there are fewer optima.  (The well tuned SmeLU slightly outperforms the well tuned Swish.)
We also observe trade-offs as function of $\beta$ for both activations consistent with those described for real and benchmark data-sets in \cite{shamir20s}.  (Similar results for SmeLU and Swish were also observed with the narrow and wide models.)  

Finally, Fig.~\ref{fig:quadratic_towers} studies the effect of \emph{warm starting} with transfer learning (TL) on the ReLU models.  Here, we trained a ReLU model with the same data generation, and used the resulting model weights to initialize the model pair, which are now trained with smaller learning rate. The lower learning rate reduces PD levels even with aggressive shuffling, maintaining competitive excess loss.   We should emphasize that the loss comparison is not a fair one, as the TL models have effectively seen twice the amount of training examples.

\section{Conclusions}
We empirically studied irreproducibility in deep networks, showing that the phenomenon exists even for the simplest possible models with the simplest possible data generation.
We demonstrated that irreproducibility can emerge from randomness in initialization of model parameters, randomness stemming from intentional (or unintentional) shuffling of the training examples. It can then be exacerbated by rounding errors, non-smooth activations, model complexity, data model, choice of model hyper-parameters and other related factors.  Smooth activations were shown to mitigate the problem under limited data shuffling, but their benefits diminish with more aggressive shuffling.  We observed how irreproducibility is manifested in the internal representations of a model, with parameters deviating from one another for comparable pairs of models.


\bibliography{repro}

\begin{thebibliography}{20}
\providecommand{\natexlab}[1]{#1}
\providecommand{\url}[1]{\texttt{#1}}
\expandafter\ifx\csname urlstyle\endcsname\relax
  \providecommand{\doi}[1]{doi: #1}\else
  \providecommand{\doi}{doi: \begingroup \urlstyle{rm}\Url}\fi

\bibitem[Achille et~al.(2017)Achille, Rovere, and Soatto]{achille17}
Achille, A., Rovere, M., and Soatto, S.
\newblock Critical learning periods in deep neural networks.
\newblock \emph{arXiv preprint arXiv:1711.08856}, 2017.

\bibitem[Anil et~al.(2018)Anil, Pereyra, Passos, Ormandi, Dahl, and
  Hinton]{anil18}
Anil, R., Pereyra, G., Passos, A., Ormandi, R., Dahl, G.~E., and Hinton, G.~E.
\newblock Large scale distributed neural network training through online
  distillation.
\newblock \emph{arXiv preprint arXiv:1804.03235}, 2018.

\bibitem[Bengio et~al.(2009)Bengio, Louradour, Collobert, and Weston]{bengio09}
Bengio, Y., Louradour, J., Collobert, R., and Weston, J.
\newblock Curriculum learning.
\newblock In \emph{Proceedings of the 26th annual international conference on
  machine learning}, pp.\  41--48, 2009.

\bibitem[Bhojanapalli et~al.(2021)Bhojanapalli, Wilber, Veit, Rawat, Kim,
  Menon, and Kumar]{bho21}
Bhojanapalli, S., Wilber, K.~J., Veit, A., Rawat, A.~S., Kim, S., Menon, A.~K.,
  and Kumar, S.
\newblock On the reproducibility of neural network predictions, 2021.

\bibitem[Chen et~al.(2020)Chen, Wang, Lin, Cheng, Hong, Chi, and Cui]{chen20}
Chen, Z., Wang, Y., Lin, D., Cheng, D., Hong, L., Chi, E., and Cui, C.
\newblock Beyond point estimate: Inferring ensemble prediction variation from
  neuron activation strength in recommender systems.
\newblock \emph{arXiv preprint arXiv:2008.07032}, 2020.

\bibitem[D'Amour et~al.(2020)D'Amour, Heller, Moldovan, Adlam, Alipanahi,
  Beutel, Chen, Deaton, Eisenstein, Hoffman, Hormozdiari, Houlsby, Hou, Jerfel,
  Karthikesalingam, Lucic, Ma, McLean, Mincu, Mitani, Montanari, Nado,
  Natarajan, Nielson, Osborne, Raman, Ramasamy, Sayres, Schrouff, Seneviratne,
  Sequeira, Suresh, Veitch, Vladymyrov, Wang, Webster, Yadlowsky, Yun, Zhai,
  and Sculley]{damour20}
D'Amour, A., Heller, K., Moldovan, D., Adlam, B., Alipanahi, B., Beutel, A.,
  Chen, C., Deaton, J., Eisenstein, J., Hoffman, M.~D., Hormozdiari, F.,
  Houlsby, N., Hou, S., Jerfel, G., Karthikesalingam, A., Lucic, M., Ma, Y.,
  McLean, C., Mincu, D., Mitani, A., Montanari, A., Nado, Z., Natarajan, V.,
  Nielson, C., Osborne, T.~F., Raman, R., Ramasamy, K., Sayres, R., Schrouff,
  J., Seneviratne, M., Sequeira, S., Suresh, H., Veitch, V., Vladymyrov, M.,
  Wang, X., Webster, K., Yadlowsky, S., Yun, T., Zhai, X., and Sculley, D.
\newblock Underspecification presents challenges for credibility in modern
  machine learning, 2020.

\bibitem[Dietterich(2000)]{dietterich00}
Dietterich, T.~G.
\newblock Ensemble methods in machine learning.
\newblock \emph{Lecture Notes in Computer Science}, pp.\  1--15, 2000.

\bibitem[Duchi et~al.(2011)Duchi, Hazan, and Singer]{duchi11}
Duchi, J., Hazan, E., and Singer, Y.
\newblock Adaptive subgradient methods for online learning and stochastic
  optimization.
\newblock \emph{Journal of Machine Learning Research}, 12:\penalty0 2121--2159,
  Feb. 2011.

\bibitem[Dusenberry et~al.(2020)Dusenberry, Tran, Choi, Kemp, Nixon, Jerfel,
  Heller, and Dai]{dusenberry20}
Dusenberry, M.~W., Tran, D., Choi, E., Kemp, J., Nixon, J., Jerfel, G., Heller,
  K., and Dai, A.~M.
\newblock Analyzing the role of model uncertainty for electronic health
  records.
\newblock In \emph{Proceedings of the ACM Conference on Health, Inference, and
  Learning}, pp.\  204--213, 2020.

\bibitem[Hinton et~al.(2015)Hinton, Vinyals, and Dean]{hinton15}
Hinton, G., Vinyals, O., and Dean, J.
\newblock Distilling the knowledge in a neural network.
\newblock \emph{arXiv preprint arXiv:1503.02531}, 2015.

\bibitem[Lakshminarayanan et~al.(2017)Lakshminarayanan, Pritzel, and
  Blundell]{lakshminarayanan17}
Lakshminarayanan, B., Pritzel, A., and Blundell, C.
\newblock Simple and scalable predictive uncertainty estimation using deep
  ensembles.
\newblock In \emph{Advances in neural information processing systems}, pp.\
  6402--6413, 2017.

\bibitem[McMahan et~al.(2013)McMahan, Holt, Sculley, Young, Ebner, Grady, Nie,
  Phillips, Davydov, Golovin, et~al.]{mcmahan13}
McMahan, H.~B., Holt, G., Sculley, D., Young, M., Ebner, D., Grady, J., Nie,
  L., Phillips, T., Davydov, E., Golovin, D., et~al.
\newblock Ad click prediction: a view from the trenches.
\newblock In \emph{Proceedings of the 19th ACM SIGKDD international conference
  on Knowledge discovery and data mining}, pp.\  1222--1230, 2013.

\bibitem[Nagarajan et~al.(2018)Nagarajan, Warnell, and Stone]{nagarajan18}
Nagarajan, P., Warnell, G., and Stone, P.
\newblock Deterministic implementations for reproducibility in deep
  reinforcement learning.
\newblock \emph{arXiv preprint arXiv:1809.05676}, 2018.

\bibitem[Nair \& Hinton(2010)Nair and Hinton]{nair10}
Nair, V. and Hinton, G.~E.
\newblock Rectified linear units improve restricted boltzmann machines.
\newblock In \emph{ICML}, 2010.

\bibitem[Ramachandran et~al.(2017)Ramachandran, Zoph, and Le]{ramachandran17}
Ramachandran, P., Zoph, B., and Le, Q.~V.
\newblock Searching for activation functions.
\newblock \emph{arXiv preprint arXiv:1710.05941}, 2017.

\bibitem[Shamir(2018)]{shamir18}
Shamir, G.~I.
\newblock Systems and methods for improved generalization, reproducibility, and
  stabilization of neural networks via error control code constraints, 2018.

\bibitem[Shamir \& Coviello(2020)Shamir and Coviello]{shamir20a}
Shamir, G.~I. and Coviello, L.
\newblock Anti-distillation: Improving reproducibility of deep networks.
\newblock \emph{arXiv preprint arXiv:2010.09923}, 2020.

\bibitem[Shamir et~al.(2020)Shamir, Lin, and Coviello]{shamir20s}
Shamir, G.~I., Lin, D., and Coviello, L.
\newblock Smooth activations and reproducibility in deep networks.
\newblock \emph{arXiv preprint arXiv:2010.09931}, 2020.

\bibitem[Summers \& Dinneen(2021)Summers and Dinneen]{summers21}
Summers, C. and Dinneen, M.~J.
\newblock On nondeterminism and instability in neural network optimization,
  2021.

\bibitem[Zhang et~al.(2018)Zhang, Xiang, Hospedales, and Lu]{zhang18}
Zhang, Y., Xiang, T., Hospedales, T.~M., and Lu, H.
\newblock Deep mutual learning.
\newblock In \emph{Proceedings of the IEEE Conference on Computer Vision and
  Pattern Recognition}, pp.\  4320--4328, 2018.

\end{thebibliography}
\bibliographystyle{icml2021}



\end{document}